# Three-dimensional reconstruction and segmentation of an aggregate stockpile for size and shape analyses

## Reconstruction tridimensionnelle et segmentation d'un stock d'agrégats pour les analyses de taille et de forme

**Erol Tutumluer**, Haohang Huang, Jiayi Luo & Issam Qamhia
*Department of Civil and Environmental Engineering, University of Illinois at Urbana-Champaign, USA*
tutumlue@illinois.edu

John M. Hart
*Beckman Institute for Advanced Science and Technology, University of Illinois at Urbana-Champaign, USA*

ABSTRACT: Aggregate size and shape are key properties for determining quality of aggregate materials used in road construction and transportation geotechnics applications. The composition and packing, layer stiffness, and load response are all influenced by these morphological characteristics of aggregates. Many aggregate imaging systems developed to date only focus on analyses of individual or manually separated aggregate particles. There is a need to develop a convenient and affordable system for acquiring 3D aggregate information from stockpiles in the field. This paper presents an innovative 3D imaging approach for potential field evaluation of large-sized aggregates, whereby engineers can perform inspection by taking videos/images with mobile devices such as smartphone cameras. The approach leverages Structure-from-Motion (SfM) techniques to reconstruct the stockpile surface as 3D spatial data, i.e. point cloud, and uses a 3D segmentation algorithm to separate and extract individual aggregates from the reconstructed stockpile. The preliminary results presented in this paper demonstrate the future potential of using 3D aggregate size and shape information for onsite Quality Assurance/Quality Control (QA/QC) tasks.

RÉSUMÉ : Les propriétés de taille et de forme sont des propriétés clés pour déterminer la qualité des granulats utilisés dans la construction routière et géotechnique des transports. La composition et le tassement, la rigidité de la couche et la réponse à la charge sont tous influencés par les caractéristiques morphologiques des granulats. Les nombreux systèmes d'imagerie de granulats développés à ce jour se concentrent uniquement sur les analyses des particules individuelles ou séparées manuellement. Il est donc nécessaire de développer un système pratique et abordable pour acquérir de l'information sur les particules en 3D directement à partir des piles en chantier. Cet article présente une approche innovante d'imagerie 3D pour une évaluation de granulats de grande taille sur le terrain, où les ingénieurs peuvent effectuer une inspection en prenant des vidéos / images directement de leurs appareils mobiles avec caméra. L'approche exploite les techniques de structure à partir du mouvement (Structure-from-Motion) pour reconstruire et modéliser en 3D la surface de la pile, i.e. nuage de points, et utilise un algorithme de segmentation 3D pour séparer et extraire les particules de la pile. Les résultats préliminaires présentés dans cet article démontrent le potentiel d'utiliser les informations 3D des granulats pour l'assurance qualité / contrôle qualité (AQ / CQ) sur le terrain.

## 1 INTRODUCTION

Riprap and large-sized aggregates are being widely used in geotechnical and hydraulic applications. At the production sites, crushed stone aggregate producers first perform quarrying to excavate rock from the ground and crush them into large-sized particles, which can then be screened into specific sizes for immediate use or further processing. Aggregate materials at quarry sites are typically categorized by size and stored in separate stockpiles. Size and morphological properties of aggregates are primary factors that influence the performance of aggregate materials in various transportation infrastructure applications such as Portland cement concrete surface (Quiroga et al. 2004), unbound or bound aggregate base/subbase (Tutumluer & Pan 2008), ballast in railway tracks (Wnek et al. 2013), and riprap materials for slope stability and erosion control in hydraulic applications (Lagasse et al. 2006). Information on the size and shape properties of aggregates in a stockpile can facilitate the Quality Assurance/ Quality Control (QA/QC) process and help understand the material behavior (Barrett 1980).

According to Lagasse et al. (2006), standardized specifications or guidelines that ensure reliable and efficient characterization of weight, size, shape, and gradation of riprap categories are critical at both production lines and construction sites. A nationwide American Association of State Highway and Transportation Officials (AASHTO) survey of transportation agencies in the US and Canada has indicated that the current practice of riprap characterization is mostly based on visual inspection and/or manual size and weight measurements (Sillick 2017). Visual inspection provides a rough size estimate of the aggregates in a stockpile but is subjective and uncertain. Further, manual size measurements require heavy machinery to handle individual rocks, making the process time-consuming and labor-intensive. Both methods are qualitative in terms of shape characterization, and both lack the capability to capture the morphological properties of aggregates. As a result, a major challenge for characterizing the size and shape properties of riprap is primarily due to the difficulties associated with its large size and heavy weight.

In this regard, aggregate imaging techniques, which have been developed over the past two decades, may provide a solution for the quantitative analysis of aggregate materials (Rao et al. 2002, Al-Rousan et al. 2005, Pan et al. 2006, Wang et al. 2013, Moaveni et al. 2013, and Hryciw et al. 2014). Most of the current aggregate imaging techniques follow the procedure: (a) individual aggregate particles are manually arranged in a laboratory setup under well-controlled background and lighting conditions, (b) a camera system captures the image of aggregates, and (c) a computer program analyzes the images to obtain the size and shape properties. However, these techniques are only applicable to small and medium-sized aggregates that can be easily handled in a laboratory setup and are therefore not scalable for characterizing large-sized aggregates. The in-place inspection at production or construction sites also poses additional challenges to the image acquisition and analysis steps due to the natural background and the lighting conditions. To address some of these challenges, a field imaging approach for large-sized aggregates has been recently developed to analyze the stockpile aggregate information from 2D images (Huang et al. 2019, Huang et al. 2020).

As mentioned, two decades of recent developments in aggregate imaging provided efficient tools for capturing the aggregate properties from 2D images, but 3D analysis of aggregate stockpiles is still challenging and seldomly conducted. There are certain limitations of 2D imaging approaches since significant amount of useful spatial information is lost when projecting a 3D scene onto a 2D image plane. 3D size and shape information can offer more comprehensive geometric features as well as more accurate characterization of riprap and large-sized material. Most of the previous work focused on the scanning of individual aggregates using 3D laser scanners, X-ray CT scanners, or depth cameras (Komba et al. 2013, Anochie-Boateng et al. 2013, Zheng & Hryciw 2014, Jin et al. 2018). A convenient and practical approach for performing 3D analysis of in-place aggregate stockpiles has not yet been developed.

This paper focuses on the development of reliable and efficient 3D imaging techniques that can facilitate convenient QA/QC checks for accurately evaluating aggregate stockpiles. Through 3D reconstruction and segmentation analyses of aggregate stockpiles, improved characterizations of size and shape properties and optimized material selection in the field can be achieved to improve designs through effective quality control, reduced costs, increased life cycle, and minimum labor and energy consumption. Major cost savings in terms of personnel time, transportation, and laboratory equipment and facility use can be realized.

## 2 OBJECTIVE AND METHODOLOGY

The objective of this research is to investigate the relevant computer vision techniques and develop a preliminary 3D imaging approach for aggregate stockpiles. The imaging approach is intended for potential field evaluation of large-sized aggregates, whereby engineers can perform inspection by taking videos/images with mobile devices such as smartphone cameras. The proposed approach leverages Structure-from-Motion (SfM) techniques to reconstruct the stockpile surface as 3D spatial data, i.e. point cloud, and develops a 3D segmentation algorithm to separate and extract individual aggregates from the reconstructed stockpile. The results on the reconstruction and segmentation can serve as a proof-of-concept for the potential of incorporating 3D aggregates information during onsite QA/QC tasks.

## 3 3D IMAGING APPROACH FOR AGGREGATE STOCKPILES

### 3.1 *3D Reconstruction of Stockpile Surface as Point Cloud*

To overcome the limitations of single-view 2D image representation of a stockpile, a 3D reconstruction approach is proposed herein. A sequence of images taken from multiple views or frames extracted from a video captured with a moving camera are used as the input in the 3D reconstruction step. An ideal set of images of a stockpile should capture views covering multiple heights and viewing angles, which can be obtained by moving the camera around the stockpile surface.

In the computer vision domain, Structure-from-Motion (SfM) techniques are widely used for the 3D reconstruction tasks. SfM refers to the recovering of 3D stationary structure from a collection of 2D images. SfM involves three main stages: (a) extracting local features from 2D images and matching these features across images, (b) estimating the motion of cameras and obtaining relative camera translation and rotation information, and (c) recovering the 3D structure by jointly using the estimated camera motion information and image features and minimizing the total reprojection error. The fundamentals and detailed formulations of the 3D SfM approach were first proposed by Longuet-Higgins (1981) and then summarized by Hartley and Zisserman (2003).

In computational photography, 3D points are projected to pixels in a 2D image via the projection matrix, $P$, which is a $3 \times 4$ matrix that encodes the translation, orientation, and internal calibration parameters of the camera. Using homogeneous coordinates, a 3D point $\left\{\frac{X}{W}, \frac{Y}{W}, \frac{Z}{W}, 1\right\}$ is projected to a 2D pixel $\left\{\frac{x}{z}, \frac{y}{z}, 1\right\}$ by multiplying with the projection matrix. For a collection of 2D images taken from multiple views, the projection matrix at each view is different and unknown. Moreover, the coordinates of the 3D point are also unknown. The only available information in 3D reconstruction is the pixel correspondences among different views. As shown in Figure 1, if a 3D point is simultaneously visible in multiple views, the point coordinates and the camera parameters can be jointly estimated. This process is defined as the bundle adjustment technique in 3D reconstruction research (Triggs et al. 1999).

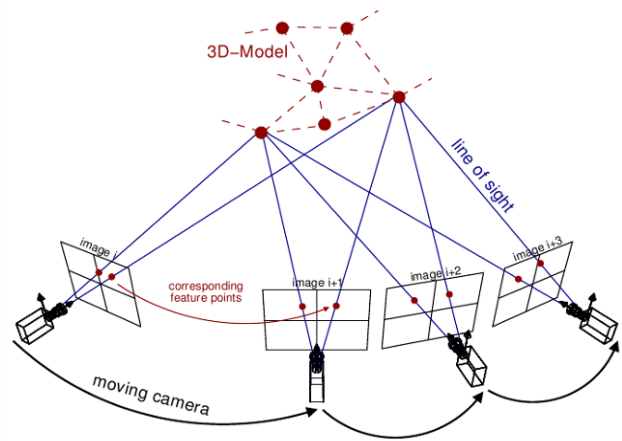

Figure 1. Structure-from-Motion (SfM) technique.

Given $m$ images of $n$ 3D points, the goal of bundle adjustment is to estimate (a) the projection matrix $P_i$ associated with each camera and (b) the coordinates $X_j$ of each 3D point. This is a minimization problem on the cost function called the total reprojection error (see Eq. 1).

$$\underset{\{P,X\}}{\text{minimize}} \sum_{i=1}^{m}\sum_{j=1}^{n}\left\|x_{ij} - P_i X_j\right\|^2 \quad (1)$$

where $x_{ij}$ is the projected pixel of point $X_j$ in $i$-th camera view, $P_i$ is the projection matrix of $i$-th camera, and $X_j$ is the coordinates of $j$-th point. The total reprojection error is the squared pixel distance of all feature points among all camera views. Bundle adjustment minimizes the reprojection error by iteratively finding the best estimates of the camera parameters and the point coordinates. After convergence, the estimated camera translation and orientation of each image are localized in the space, and the reconstructed scene structure is represented as 3D point cloud data. Point cloud data consist of points with 3D spatial coordinates, associated with the RGB color values and surface normal directions at the points. Based on the SfM techniques, images need to be taken from multiple views around the stockpile to reconstruct it as a 3D point cloud. The surfaces of the aggregates in the stockpile are represented as points with different colors and facings.

### 3.2 *Point Cloud Segmentation for Individual Aggregates*

The reconstructed 3D point cloud accumulates and provides more comprehensive spatial information of an aggregate stockpile than in the case of single-view 2D images. However, the size and shape information of individual riprap particles are not directly accessible from the raw reconstructed point cloud. Therefore, a 3D point cloud segmentation approach is required to obtain useful morphological information of individual riprap particles in the stockpile.

First, the 3D point cloud data is transformed into 3D mesh data. This is because the point cloud format is known as an irregular and unordered representation. By transforming to mesh data, more efficient mesh segmentation analyses can be performed. For this purpose, Poisson surface reconstruction technique (Kazhdan et al. 2006) was used to form a triangular mesh representation from the point cloud. Poisson surface reconstruction applies on oriented points (points with normal directions) and form a surface that conforms to the normal direction at each point. This step is obtained by solving the Poisson's partial differential equation. After the Poisson surface reconstruction, the stockpile surface is represented as a triangular mesh, with "vertices" being the points and "faces" indicating the connectivity among the points.

A curvature-based segmentation algorithm is then developed in this preliminary study for separating individual aggregates in the mesh of the reconstructed stockpile. The segmentation algorithm is a customized Breadth-First Search (BFS) algorithm with curvature constraints during the search process. BFS algorithms are commonly applied to tree or graph data structures and are widely used in computer science and graph research in mathematics (Knuth 1997). The typical procedure of BFS on meshes is to start from an initial face, add the neighboring faces to a queue data structure, and iteratively traverse the entire mesh.

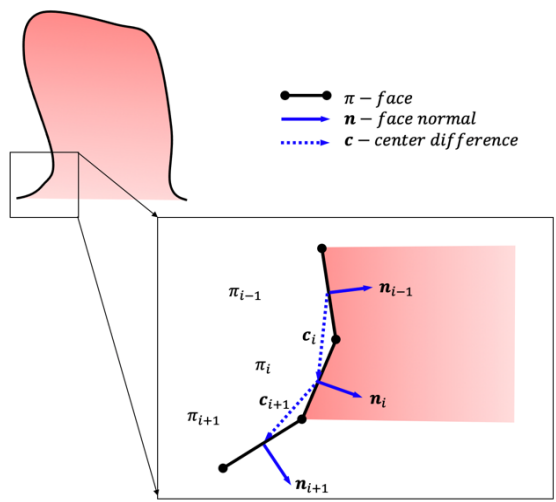

(a)

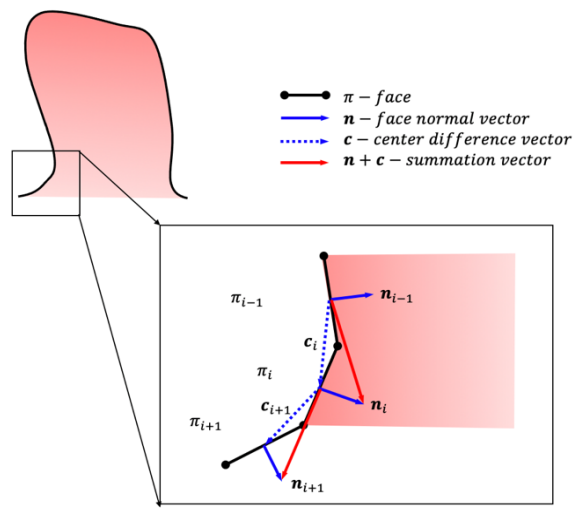

(b)

Figure 2. (a) Face normal vector and center difference vector, and (b) curvature-based criterion.

For the mesh segmentation purpose, curvature constraints are added during the BFS procedure to identify the boundaries between individual aggregates. Details of the curvature constraints are presented in Figure 2. The faces in the mesh are denoted as $\pi$, where $\pi_{i-1}, \pi_i, \pi_{i+1}$ in Figure 2(a) represents the adjacent faces. Since each face is a triangle that has three edges, it will have three adjacent neighboring faces. The normal vectors of the faces are denoted by $n$, and a center difference vector, $c$, is defined by subtracting the centroids of two neighboring faces. Note that although Figure 2 uses a 2D illustration for brevity, the faces are triangles in 3D space, and the normal vectors and center difference vectors are both 3D vectors. The key idea in the mesh segmentation algorithm is based on the observation that highly concave curvatures usually appear at the boundaries between aggregates, while the surface of individual aggregates typically has a convex or slightly concave curvature. Therefore, a curvature-based criterion is developed [see Eq. 2 and Figure 2(b)] to be used in the customized BFS algorithm.

$$(c_{i+1} + n_{i+1}) \cdot n_i > threshold\ t \qquad (2)$$

When the BFS is at face $\pi_i$, the algorithm checks its unvisited neighboring face $\pi_{i+1}$ and uses Eq. 2 to decide whether or not the BFS reaches a potential aggregate boundary. The equation computes the dot product between (a) the summation of center difference vector $c_{i+1}$ and normal vector $n_{i+1}$ of the neighboring face and (b) the normal vector of the current face $n_i$. The magnitude of the dot product captures the relative curvature between the two adjacent faces. When Eq. 2 is satisfied, it indicates the two adjacent faces have a relatively convex curvature, and the BFS continues by adding the neighboring face $\pi_{i+1}$ to the queue; otherwise, the BFS stops and marks the neighboring face $\pi_{i+1}$ as part of the aggregate boundaries. From experiments, the threshold value used in this preliminary study is 0.7, which corresponds to an approximately 45° angle between the center difference vector and normal vector.

The mesh segmentation algorithm runs with automatic restart until the entire stockpile surface mesh is traversed. As described above, one run of the BFS starts from an initial face of an aggregate, traverses until it reaches the boundary with neighboring aggregates, and marks the boundary faces. Upon completion, the visited faces during this BFS run are collected and reported as one segmented aggregate. Then, another run of BFS will start from another unvisited face and iterates until the mesh is fully segmented.

4 RESULTS AND DISCUSSION

A publicly available software VisualSFM (Wu 2013) was used for the 3D reconstruction of a small stockpile made from 10 riprap rocks. These 10 rocks are relatively large-sized aggregates, with sizes ranging from 3 in. to 10 in. Smartphone cameras, which are commonly accessible and convenient, were used for capturing images of the stockpile. The reconstruction precision depends on the number of multi-view images as well as the image resolution. In this experiment, a total of 46 multi-view images were taken from varying heights and viewing angles covering different views of the stockpile, as shown in Figure 3(a). The image resolution of the camera was 2400-pixels by 3000-pixels, which can be achieved by most smartphones with high-resolution cameras. Other types of digital cameras can also be used if the images collected are of high quality and proper resolution. The dense 3D point cloud generated by VisualSFM is shown in Figure 3(b), with the localized camera positions shown above the point cloud.

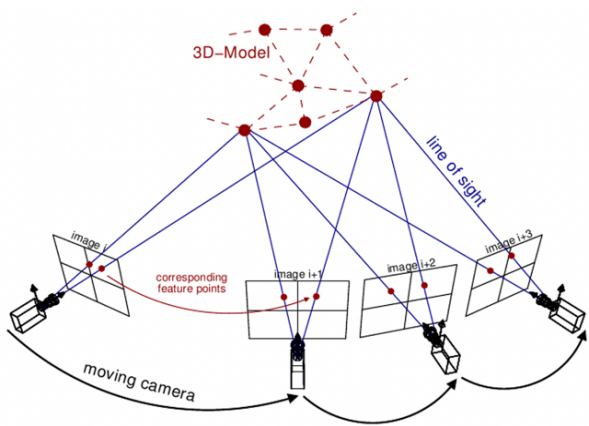

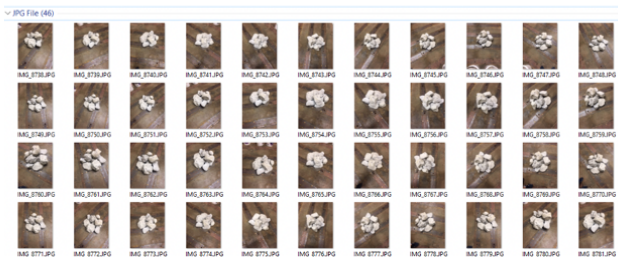

(a)

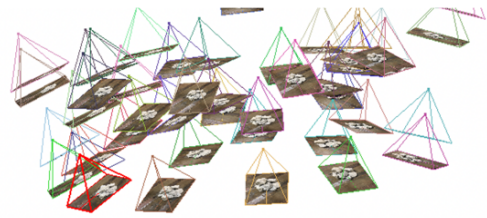

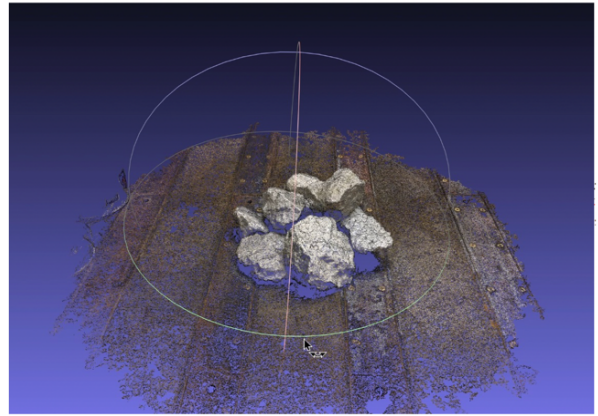

(b)

Figure 3. (a) Collection of multi-view images for 3D reconstruction and (b) 3D reconstructed point cloud of the stockpile from multi-view geometry.

A comparison by aligning one of the multi-view images and the reconstructed point cloud is also presented in Figure 4(a) and Figure 4(b). As shown, the reconstructed point cloud is visually realistic and of high fidelity, with the rock surface and boundaries appearing similar to the real scene. The surface texture, rock color, and the marker signs on the rocks are all identifiable in the point cloud, which indicates it captures and recovers the stockpile information with a high level of details. By applying the Poisson surface reconstruction and the curvature-based BFS segmentation algorithm, the stockpile surface was segmented as presented in Figure 4(c). Faces belonging to different segmented aggregates were labeled with different colors, and all 10 rocks were successfully segmented from the reconstructed stockpile. As shown in Figure 4(c), the separation of adjacent aggregates identifies the boundaries and is consistent with human perception.

The 3D reconstruction and segmentation results provide a proof-of-concept and the potential application of the 3D imaging approach for stockpile analyses. To inspect a stockpile in the field, engineers can capture multi-view images or a video using smartphone cameras by moving around the stockpile. The stockpile surface can then be reconstructed as a 3D point cloud, and individual aggregates can be extracted using the segmentation algorithm. Further, for each segmented aggregate, size and shape analyses can be conducted to obtain the 3D morphological properties of the material. The results can be further utilized to derive the gradation and typical shape patterns for a batch of aggregate products. Note that the reconstruction and segmentation results only represent properties of those aggregates on the stockpile surface since only the surface is visible for an in-place inspection. This is typically the case when manual QA/QC inspection methods are conducted.

In summary, the 3D reconstruction and segmentation approach presented in this paper demonstrates the capability of capturing aggregate stockpile information as 3D data and pre-processing the data for further morphological analyses. The 3D morphological analysis is an ongoing research effort of the authors and is beyond the scope of this paper.

Further developments are required to address several challenges of this proposed 3D imaging approach. First, the reconstructed stockpile is dimensionless from the raw reconstruction step. A calibration object will be needed to determine the scale of the reconstruction. Convenient tools need to be developed to enable the accurate selection and segmentation of the calibration object. Secondly, the current BFS segmentation algorithm follows a curvature-based criterion with fixed parameters. In computer vision domain, emerging techniques based on artificial intelligence and deep learning have demonstrated the potential for achieving performance matching and even surpassing human vision. The deep learning-based techniques can be utilized to improve the 3D segmentation step. Finally, since only the stockpile surface is visible for in-place inspection, the segmented aggregates have incomplete shapes. To estimate and characterize the size and shape properties of aggregates, shape prediction algorithms need to be developed to estimate the unseen side of the aggregates based on some certain expected shape, e.g. flatness and elongation, of aggregate particles. Since the typical aggregate size and shape properties can vary greatly due to different geologic origins, rock minerology and quarry crushing processes, a comprehensive database and best practice should be established to provide a priori knowledge for the shape prediction algorithms.

In addition, different reconstruction methods and devices can be considered to select the most convenient approach. For example, affordable Light Detection And Ranging (LiDAR) devices are a good alternative to acquire high-precision and calibration-free mapping for the 3D reconstruction task. Note that different 3D reconstruction approaches such as SfM and LiDAR both generate the reconstruction results in point cloud format, therefore, the segmentation approach is applicable to both cases. Moreover, both 3D reconstruction options focus on capturing the structure of objects that is less affected by environmental conditions, such as shadow effects, and hence the performance is more robust than 2D image-based techniques.

## 5 CONCLUSIONS

3D aggregate information provides more comprehensive spatial representation of aggregates than 2D images. From the literature, most of the aggregate imaging systems developed to date focused on 2D analyses of aggregate images or 3D analyses of individual aggregates. A 3D imaging approach is introduced in this paper for the in-place evaluation of aggregate stockpiles in the field. From multi-view images/videos collected by moving around the aggregate stockpile, the approach leverages Structure-from-Motion (SfM) techniques to reconstruct the stockpile surface as 3D point cloud data, and develops a 3D curvature-based segmentation algorithm to separate and extract individual aggregate particles from the reconstructed stockpile. A small stockpile was used to test and demonstrate the 3D imaging approach. The reconstruction and segmentation results show a realistic representation of the stockpile and an effective segmentation for individual aggregates. The results indicate the potential application of this convenient 3D imaging approach for aggregate stockpile analysis, whereby engineers can eventually obtain the in-place size and shape characteristics of the stockpile for onsite QA/QC checks.

## 6 ACKNOWLEDGEMENTS

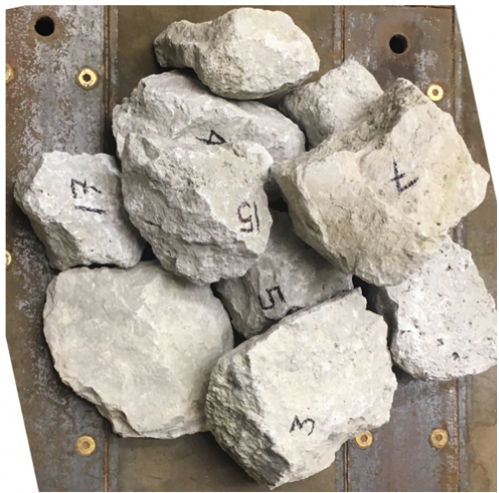

(a)

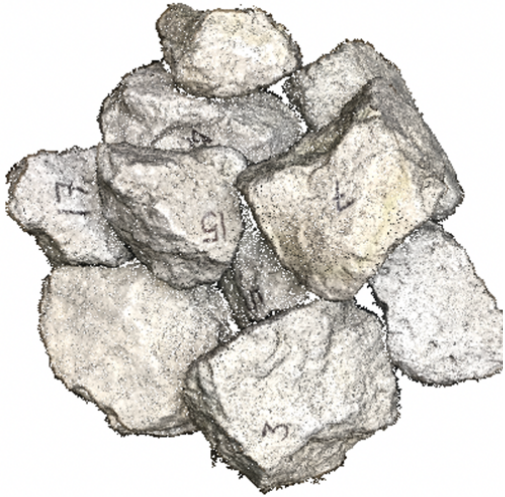

(b)

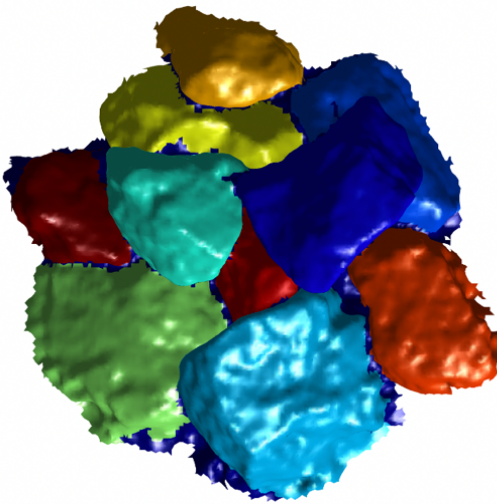

(c)

Figure 4. (a) Original stockpile scene, (b) point cloud reconstructed from SfM, and (c) 3D segmentation results for detecting individual riprap aggregates.

cooperation with the Illinois Department of Transportation (IDOT), Office of Program Development and the U.S. Department of Transportation, Federal Highway Administration. The authors would like to acknowledge the members of IDOT Technical Review Panel (TRP) for their useful advice at different stages of this research. Especially, the help and support of TRP chair Andrew Stolba with IDOT, Sheila Beshears of RiverStone Group, and Andrew Buck and Daniel Barnstable of Vulcan Materials with the data collection are greatly appreciated. The contents of this paper reflect the views of the authors who are responsible for the facts and the accuracy of the data presented. This paper does not constitute a standard, specification, or regulation.

## 7 REFERENCES

Al-Rousan, T., Masad, E., Myers, L., and C. Speigelman. New Methodology for Shape Classification of Aggregates. *Transportation Research Record: Journal of the Transportation Research Board*, No. 1913, 2005, pp. 11-23.

Anochie-Boateng, J. K., Komba, J., and G. Mvelase. Three-Dimensional Laser Scanning Technique to Quantify Aggregate and Ballast Shape Properties. *Journal of Construction and Building Materials*, Vol. 43, 2013, pp. 389-398.

Barrett, P. J. The Shape of Rock Particles, A Critical Review. *Sedimentology,* Vol. 27, 1980.

Hartley, R., and A. Zisserman. *Multiple view geometry in computer vision*. Cambridge university press, 2003.

Hryciw, R.D., Zheng, J., Ohm, H. S., and J. Li. Innovations in Optical Geo-characterization. *In Geo-Congress 2014 Keynote Lectures: Geo-Characterization and Modeling for Sustainability*, 2014, pp. 97-116. https://doi.org/10.1061/9780784413289.005.

Huang, H., Luo, J., Moaveni, M., Tutumluer, E., Hart, J. M., Beshears, S., and A. J. Stolba. Field Imaging and Volumetric Reconstruction of Riprap Rock and Large-Sized Aggregates: Algorithms and Application. *Transportation Research Record: Journal of the Transportation Research Board*, 2019. https://doi.org/10.1177/0361198119848704.

Huang, H., Luo, J., Tutumluer, E., Hart, J. M., and A. J. Stolba. Automated Segmentation and Morphological Analyses of Stockpile Aggregate Images Using Deep Convolutional Neural Networks. *Transportation Research Record: Journal of the Transportation Research Board*, 2020. https://doi.org/10.1177/0361198120943887.

Jin, C., Yang, X., You, Z., and K. Liu. Aggregate Shape Characterization Using Virtual Measurement of Three-Dimensional Solid Models Constructed from X-Ray CT Images of Aggregates. *Journal of Materials in Civil Engineering*, Vol. 30, No. 3, 2018, pp. 04018026. https://doi.org/10.1061/(ASCE)MT.1943-5533.0002210.

Komba, J., Anochie-Boateng, J., and W. van der Merwe Steyn. Analytical and Laser Scanning Techniques to Determine Shape Properties of Aggregates. *Transportation Research Record: Journal of the Transportation Research Board*, No. 2335, 2013, pp. 60-71. https://doi.org/10.3141/2335-07.

Lagasse, P. F., Clopper, P. E., Zevenbergen, L. W., and J. F. Ruff. *NCHRP Report 568: Riprap Design Criteria, Recommended Specifications, and Quality Control*. Transportation Research Board, Washington, D.C., 2006.

Longuet-Higgins, H. C. A computer algorithm for reconstructing a scene from two projections. *Nature*, 1981, 293:133–135.

Moaveni, M., Wang, S., Hart, J.M., Tutumluer, E., and N. Ahuja. Evaluation of Aggregate Size and Shape by Means of Segmentation Techniques and Aggregate Image Processing Algorithms. *Transportation Research Record: Journal of the Transportation Research Board*, No. 2335, 2013, pp. 50-59. https://doi.org/10.3141/2335-06.

Pan, T., Tutumluer, E., and J. Anochie-Boateng. Aggregate Morphology Affecting Resilient Behavior of Unbound Granular Materials. *Transportation Research Record: Journal of Transportation Research Record*, No. 1952, 2006, pp. 12-20. https://doi.org/10.3141/1952-02.

Quiroga, P. N., and D. W. Fowler. *ICAR Research Report No. 104-1F: The Effects of Aggregate Characteristics on The Performance of Portland Cement Concrete.* International Center for Aggregates Research, 2004.

Rao, C., Tutumluer, E., and I.T. Kim. Quantification of Coarse Aggregate Angularity based on Image Analysis. *Transportation Research Record: Journal of the Transportation Research Board*, No. 1787, 2002, pp. 117-124. https://doi.org/10.3141/1787-13.

Sillick, S. Montana Department of Transportation. AASHTO RAC Member Survey Results, July 2017.
https://research.transportation.org/rac-survey-detail/?survey_id=371
Accessed July 20, 2018.

Triggs, B., McLauchlan, P. F., Hartley, R., and A. W. Fitzgibbon. Bundle adjustment—a modern synthesis. In *International workshop on vision algorithms*, pp. 298-372. Springer, Berlin, Heidelberg, 1999.

Tutumluer, E., and T. Pan. Aggregate Morphology Affecting Strength and Permanent Deformation Behavior of Unbound Aggregate Materials. *Journal of Materials in Civil Engineering*, Vol. 20, No. 9, 2008, pp. 617-627.

Wang, L., Sun, W., Tutumluer, E., and C. Druta. Evaluation of Aggregate Imaging Techniques for Quantification of Morphological Characteristics. *Transportation Research Record: Journal of the Transportation Research Board*, No. 2335, 2013, pp. 39-49. https://doi.org/10.3141/2335-05.

Wnek, M. A., Tutumluer, E., Moaveni, M., and E. Gehringer. Investigation of Aggregate Properties Influencing Railroad Ballast Performance. *Transportation Research Record: Journal of Transportation Research Board*, 2013, 2374: 180-190.

Wu, C. Towards linear-time incremental structure from motion. In *2013 International Conference on 3D Vision-3DV*, pp. 127-134. IEEE, 2013.

Zheng, J., and R. D. Hryciw. Soil Particle Size Characterization by Stereo-photography. *In Geo-Congress 2014: Geo-characterization and Modeling for Sustainability*, 2014, pp. 64-73. https://doi.org/10.1061/9780784413272.007.

Kazhdan, M., Bolitho, M., and H. Hoppe. Poisson surface reconstruction. In *Proceedings of the 4th Eurographics symposium on Geometry processing*, Vol. 7. 2006.

Knuth, D. E. *The art of computer programming*. Vol. 3. Pearson Education, 1997.